\documentclass[preprint,12pt]{elsarticle}

\usepackage{amssymb}

\usepackage{lineno,hyperref}

\usepackage{graphicx}
\usepackage[outdir=./images/]{epstopdf}

\usepackage{amsmath}
\usepackage[ruled,linesnumbered]{algorithm2e}

\newtheorem{definition}{Definition}[section]

\usepackage{booktabs}

\usepackage{url}

\newcommand{\cAlgorithmName}{ACXON}

\usepackage{xcolor}

\journal{Information Sciences}

\date{February 21, 2026}

\makeatletter
\def\ps@pprintTitle{%
 \let\@oddhead\@empty
 \let\@evenhead\@empty
 \def\@oddfoot{\footnotesize\itshape
       Accepted in \@journal. DOI: https://doi.org/10.1016/j.ins.2026.123280.} 
 \let\@evenfoot\@oddfoot}
\makeatother

\begin{document}

\begin{frontmatter}



\title{Ontological Foundations for Contrastive Explanatory Narration of Robot Plans}


\author[iriAffiliation]{Alberto Olivares-Alarcos\corref{correspondingauthor}}
\ead[url]{aolivares@iri.upc.edu}

\author[iriAffiliation]{Sergi Foix}

\author[iriAffiliation]{Júlia Borr\`as}

\author[kclAffiliation]{Gerard Canal}

\author[iriAffiliation]{Guillem Aleny\`a}

\affiliation[iriAffiliation]{organization={Institut de Robòtica i Informàtica Industrial, Consejo Superior de Investigaciones Científicas (CSIC) - Universitat Politècnica de Catalunya (UPC)},
            addressline={Llorens i Artigas 4-6}, 
            city={Barcelona},
            postcode={08028}, 
            country={Spain}}
            
\affiliation[kclAffiliation]{organization={Department of Informatics, King’s College London},
            city={London},
            country={United Kingdom}}

\begin{abstract}
Mutual understanding of artificial agents' decisions is key to ensuring a trustworthy and successful human-robot interaction. Hence, robots are expected to make reasonable decisions and communicate them to humans when needed. 
In this article, the focus is on an approach to modeling and reasoning about the comparison of two competing plans, so that robots can later explain the divergent result. First, a novel ontological model is proposed to formalize and reason about the differences between competing plans, enabling the classification of the most appropriate one (e.g., the shortest, the safest, the closest to human preferences, etc.). This work also investigates the limitations of a baseline algorithm for ontology-based explanatory narration. To address these limitations, a novel algorithm is presented, leveraging divergent knowledge between plans and facilitating the construction of contrastive narratives. Through empirical evaluation, it is observed that the explanations excel beyond the baseline method.
\end{abstract}

\begin{graphicalabstract}
\begin{figure}[h!]
  \centering
  \includegraphics[width=1.0\linewidth]{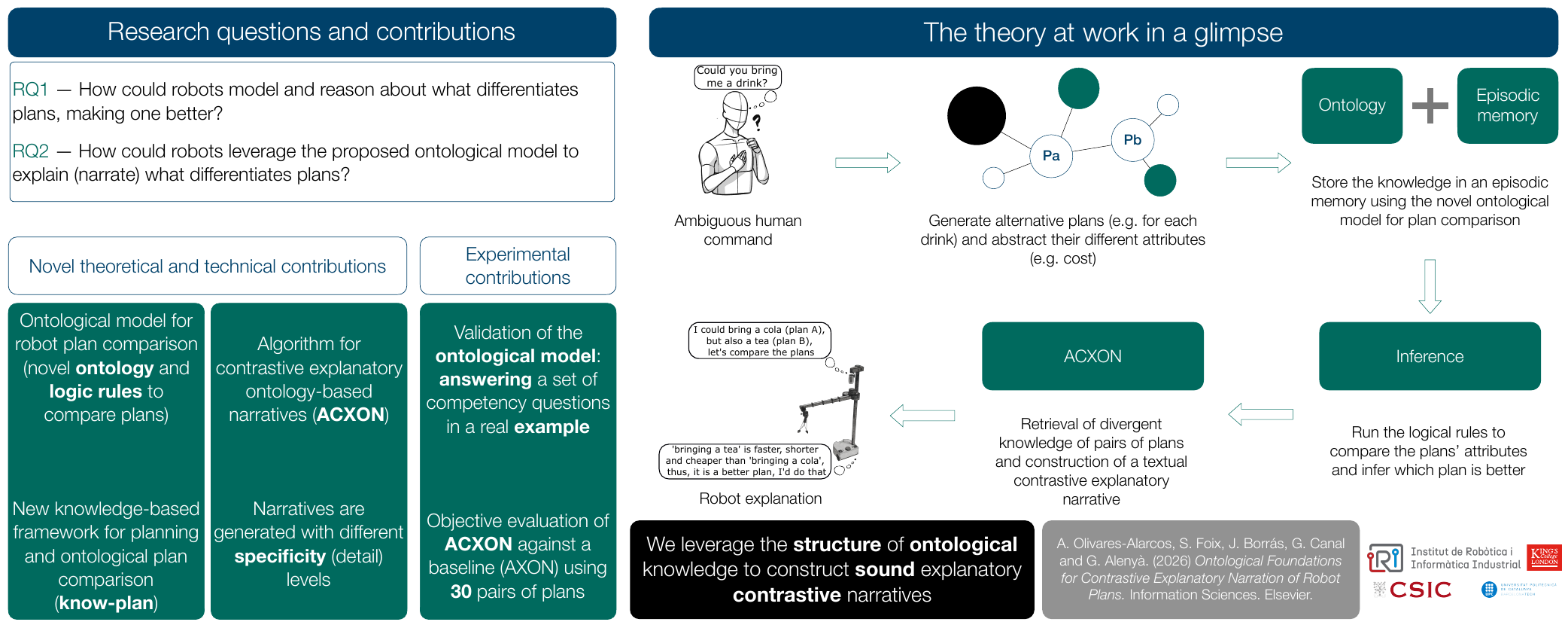}
  \label{fg:graphical_abstract}
\end{figure}
\end{graphicalabstract}

\begin{highlights}
\item A novel ontological model for plan comparison is proposed and formalized 
\item The novel ontological model is integrated with a generic planning system for robots, facilitating its re-usability
\item A set of logical rules (ontological axioms) is proposed to automate the reasoning behind the comparison of plans
\item A novel algorithm is presented to construct contrastive explanatory narratives exploiting the ontology's structure
\item The proposal is evaluated against a baseline method using several planning domains 
\end{highlights}


\end{frontmatter}


\section{Introduction}
\label{sc:intro}
Autonomous artificial decision-making in environments with different agents (e.g., robots collaborating with or assisting humans) is complex to model. This is often due to the high degree of uncertainty and potential lack of communication among agents. For instance, robots might need to choose between competing plans (i.e. sequences of actions that would allow the robot to achieve goals), comparing their properties and deciding which one is better. Note that this decision-making problem is different from finding a single plan through automated planning, as here the idea is that there are already two valid plans to execute and the robot shall compare them and identify the best one. 
This might happen when a human gives an ambiguous command (e.g. `can you bring me a drink?'), thus the robot may decompose the abstract command into different concrete goals~\cite{Izquierdo-Badiola_icra2024}, and find a plan to achieve each of the goals (such as bringing any of the available drinks). Then it would be needed to compare and disambiguate the plans. In these cases, mutual understanding of the ongoing decisions and communication between agents become crucial~\cite{doi:10.1126/scirobotics.abm4183}. Hence, trustworthy robots shall be able to model their plans' properties to make sound decisions when contrasting them. Furthermore, they shall also be capable of narrating (explaining) the knowledge acquired from the comparison. Note that robots add the possibility of physically executing the plan, which may affect the human, strongly motivating the need for explanations. This may serve two purposes: justifying the robot's selection of a plan, or asking the human to help in the disambiguation (i.e. the human may prefer the plan that the robot inferred as worse). Reflecting on these thoughts, this work addresses the following research questions:
\begin{itemize}
    \item \textbf{RQ1 -} How could robots model and reason about what differentiates plans, making one better? 
    \item \textbf{RQ2 -} How could robots leverage the proposed ontological model to explain (narrate) what differentiates plans?  
\end{itemize} 

\begin{figure}[t!]
  \centering
  \includegraphics[width=1.0\linewidth]{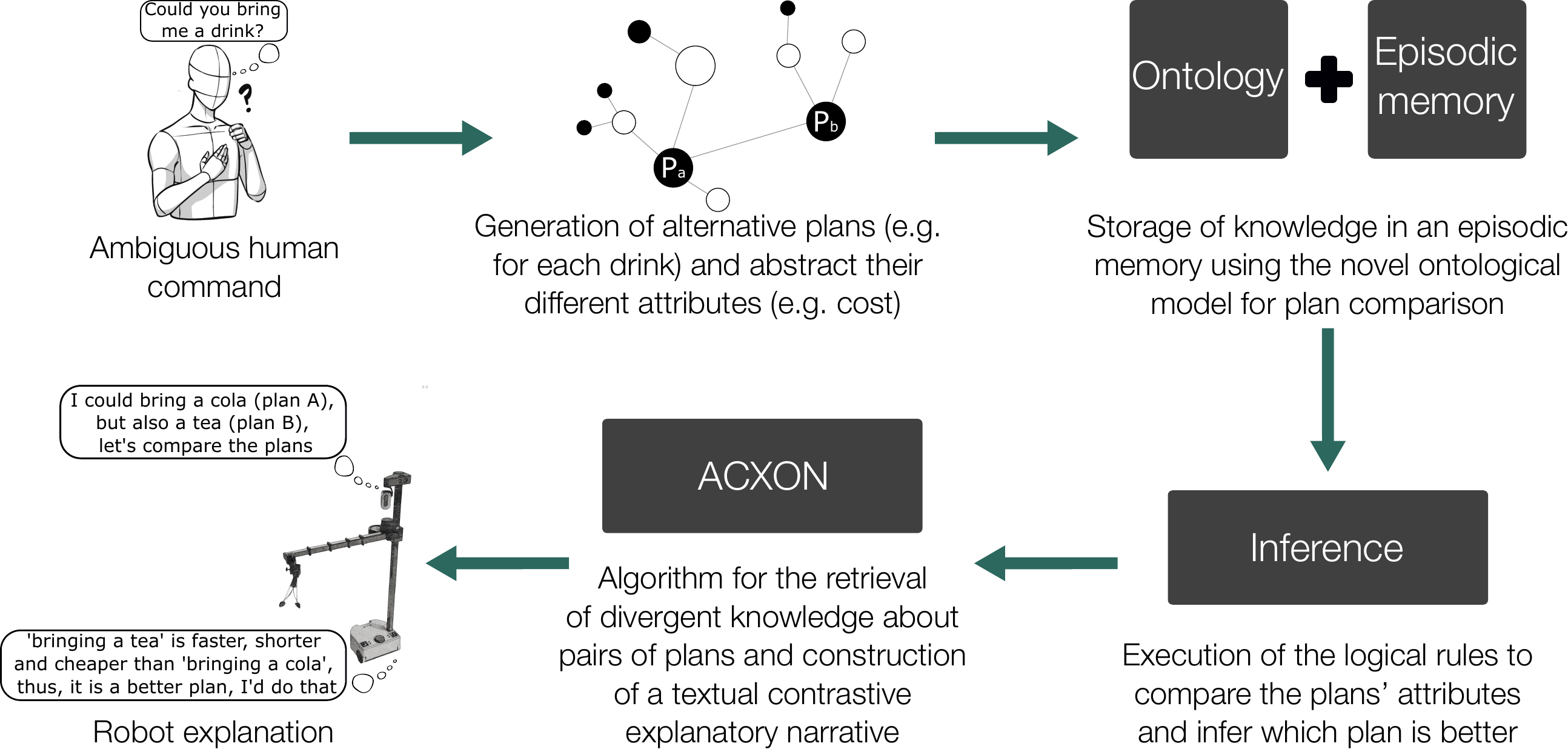}
  \caption{Proposed approach overview and its application to a prototypical human-robot interactive scenario that requires contrastive explanations. A human gives an ambiguous command, a robot generates different plans, stores knowledge about them, and contrasts them and reasons about which plan is better to execute. Finally, the inferred knowledge is used to explain the rationale.}
  \label{fg:scenario}
\end{figure}

First, an ontological analysis is conducted and a new ontological model is obtained, augmenting the scope of an ontology from the literature (OCRA~\cite{OLIVARESALARCOS2022103627}), which answers RQ1. Specifically, a new theory for plan comparison is formalized, focusing on the properties and relationships that allow comparing plans. The robot's knowledge about the plans to compare is stored, and together with some logical rules, it is used to infer which plan is better. Second, RQ2 is addressed by introducing a novel Algorithm for Contrastive eXplanatory Ontology-based Narratives ({\cAlgorithmName}), extending an existing literature methodology (XONCRA~\cite{10161359}) to contrastive cases. From the robot knowledge, {\cAlgorithmName} retrieves the divergent information about the plans, and then it constructs the final textual contrastive narrative. The proposed algorithm produces different types of narratives based on the chosen amount of detail (specificity), addressing different users' preferences. Based on objective evaluation metrics and using several planning domains, the algorithm is evaluated with respect to the original algorithm proposed in XONCRA, which is used as a baseline. The proposed algorithm outperforms the baseline, using less knowledge to build the narratives (skipping repetitive knowledge), which shortens the time to communicate the narratives. Figure~\ref{fg:scenario} provides an overview of the complete approach. At the end of the article, it is briefly discussed how the proposed algorithm can be slightly modified to enhance and restrict the knowledge selection, which helps to shorten the constructed narratives.\footnote{An extended abstract of this paper appears in the Proc. of AAMAS’24~\cite{aamas2024aoa}.} \looseness=-1  

\section{Related work}
\label{sc:related_work_contrastive_narratives}

Concerning the modeling of domain knowledge for reasoning, a common approach is to use sound formalisms such as formal ontologies. The literature shows that multiple ontologies have recently been developed~\cite{olivares-alarcos_alberto_Review2019} for different robotic applications and domains. The literature also encompasses various works that have concentrated on ontologically modeling concepts related to plans and their properties. In this regard, Bermejo-Alonso~\cite{bermejo2018reviewing} conducted a survey of that domain, reviewing existing task and planning vocabularies, taxonomies, and ontologies, while also discussing their potential integration. The surveyed works mostly defined general concepts regarding plans, and only in some cases plan properties were discussed (e.g. cost and constraints). In another work, Moulouel et al.~\cite{MOULOUEL2023468}, proposed an ontology to model contextual knowledge in partially observable environments, which is used for probabilistic commonsense reasoning and probabilistic planning. They defined concepts to model the relationship between the context in which robot actions and observations take place and other elements used during the planning (e.g. reward function).  These works are relevant in the domain, but if one wants to contrast plans during robots' decision making, there is still a need for an ontological theory that formalizes the properties of plans.  

A significant source of inspiration was discovered surrounding the notion of explainable agency (i.e., explaining the reasoning of goal-driven agents and robots). In their work, Langley et al.~\cite{10.5555/3297863.3297887} advocated for the idea that explainable agency demands four distinct functional abilities. Among those, one can find the ability to explain decisions made during plan generation, involving the comparison of alternatives. Related to this, there are some literature efforts towards investigating how to boost explainable agency by narrating or verbalizing robots' internal knowledge about plans (e.g., a plan's sequence, the rationale to include a task in the plan, etc.)~\cite{rosenthal2016verbalization,Flores2018,Canal_AAAI2022,sridharan2023}. However, it is still unexplored how robots may explain their reasoning when comparing competing plans as a whole, not just specific plan tasks.

These two research domains, ontologies and explainable agency, are certainly connected. Langley et al.~\cite{10.5555/3297863.3297887} also discussed that an explainable agent would need three main elements: a \textit{representation} of the knowledge that supports explanations, an \textit{episodic memory} to store agent experiences, and the ability to access the memory and retrieve and manipulate the stored content to \textit{construct explanations}. Olivares-Alarcos et al.~\cite{10161359}, proposed a novel methodology comprising those three elements for the construction of explanatory ontology-based narratives for collaborative robotics and adaptation (XONCRA). It consisted of a knowledge base for collaborative robotics and adaptation (\textit{know-cra}) that works as an episodic memory, and an algorithm to generate explanatory narratives using ontological knowledge (AXON). One might wonder whether the XONCRA methodology could be used to model and narrate the divergences between plans. Indeed, XONCRA uses an ontology named OCRA~\cite{OLIVARESALARCOS2022103627} that defines the relationship `is better plan than', associating two plans denoting that one of the plans is considered to be better. However, OCRA did not model how an agent might find divergences between two plans and decide which one is better. Hence, it is necessary to propose a new theory for plan comparison. Moreover, while being a potential baseline solution, the narratives generated by AXON were not optimized for a contrastive case as the one that concerns this article, and a better approach shall be investigated. 

Ontologies have already shown potential to become the integrative framework for explainable robots~\cite{10161359}. However, there are other formalisms that could complement ontologies enabling other types of reasoning. For example, non-monotonic logic supports abductive reasoning and is specifically designed to handle defeasible inferences. These allow reasoners to form tentative conclusions that can be retracted as new evidence emerges. Sridharan~\cite{sridharan2023} proposed to integrate non-monotonic logic and deep learning, exploiting their synergy across representation, reasoning, and learning layers to operationalize the principles of explainable agency. Another example of complementary formalism is causal modeling, which supports reasoning to determine the relationship between causes and their effects. Love et al.~\cite{10731403}, developed a system based on causal models to enable a robot to autonomously elicit interactions in multi-person environments, while generating real-time counterfactual explanations for its elicitation decisions. Those works presented interesting approaches to robot reasoning and explanation generation. In future work, we might consider the integration of our ontology-based methodology for explainable robots with those approaches. 

\section{Model for robot plan comparison}
\label{sc:ontology}

There exist several useful methodologies to construct ontologies, e.g., \cite{fernandez1997methontology,Spyns-TM2008Dogma}, but none arise as a definite standard. Indeed, not all those methods are suitable for this work, in which the aim is to develop an ontological model from a foundational perspective (i.e. the characterization of the main concepts is more important than the coverage of the application domain). Hence, this work relies on ontological analysis, an approach which precedes the usual ontology construction process and aims to fix the core framework for the domain ontology. Based on this selection, the steps to perform are: to set the ontology domain and scope (competency questions), to enumerate, analyze and compare existing concepts (identification of shortcomings), to develop and formalize a more solid conceptualization, and to create instances of the concepts and show their use (implementation/validation). Finally, it is also considered the documentation and maintenance of the produced theory.

\subsection{Ontological scope of the proposed theory}
The target novel model will formalize the ontological classes and relationships to represent knowledge of plans and their characteristics for plan comparison and later contrastive narration. In order to scope the subject domain to be represented in the intended model, a set of competency questions is proposed, which are a set of requirements on the ontology content. The ontological analysis started with a rough ontological scope (competency questions) focused on knowledge related to plan comparison. During the analysis, we discussed the scope until the set of questions became detailed and comprehensive enough for us to model how different plans compare. The analysis revealed that plan comparison requires modeling different characteristics of plans (CQ1), how the characteristics relate (CQ2), and how the plans relate based on their characteristics (CQ3). Specifically, the proposed ontological model is expected to be able to answer the following questions: 

\begin{itemize}
    \item \textbf{CQ1 -} Which are the characteristics of a plan? 
    \item \textbf{CQ2 -} How do the characteristics of different plans relate? 
    \item \textbf{CQ3 -} How do different plans relate to each other?  
\end{itemize}
  
The new model is going to be built upon OCRA, re-utilizing the existing model and extending it. Therefore, OCRA's upper ontology is inherited, the DOLCE+DnS Ultralite (DUL) foundational ontology~\cite{Borgo2021}. In addition to the proposed competency questions, for this work it is also interesting to represent the sequence of actions included in a plan, which is already covered by the DUL ontology. 

\subsection{Ontological shortcomings in OCRA and their theoretical remedy}
Following the ontological analysis process, this section identifies the need for new ontological concepts and relations to cover the proposed scope (i.e. the competency questions), and defines and formalizes them. The novel theory is the first in the literature for plan comparison representation and reasoning. The model is comprehensive and is built upon existing ontologies, a great practice in ontology development, and is conceived from a foundational perspective.

\subsubsection{Which are the characteristics of a plan? (CQ1)} \label{sc:modeling_cq1}
OCRA did not define any ontological classes or relationships to model the properties of plans, thus an extension is required to be able to answer CQ1. For such an extension, a top-down approach is followed and the new model is built upon general entities defined in the upper-level ontology, DUL. In order to represent the features of other entities, DUL defines the class \textit{Quality} as \textit{`any aspect of an \textit{Entity} (but not a part of it), which cannot exist without that \textit{Entity}'}. DUL also includes the relationship \textit{`has quality'}, \textit{`a relation between entities and qualities'}. In this work, we specialize both the class and the relation to define the particular qualities of plans. 

Plans can have many different qualities that would highly depend on the application domain. Defining all of them is out of the scope of this article. Instead, we aim to find a set of qualities that are usually present in most of the planning domains, with a special focus on those more relevant to robotics. Particularly, we will use temporal planning domains in which actions have a duration and are modeled using PDDL 2.1~\cite{fox2003pddl2}. In robotics, finding a plan that is valid is just the first part of the work to do, because the focus is on the execution of the plan. Hence, considering the estimated duration (or makespan) of actions makes much more sense than in other artificial intelligence domains. After carefully studying temporal planning problems, it was discovered that three major generic qualities of plans were: cost, expected makespan, and number of tasks (i.e. number of sequenced actions of a plan). The proposed definition and formalization for each of the qualities is as follows:\looseness=-1

\begin{definition}
\textit{Plan Cost is a Quality of a Plan that captures the cost of executing the Plan.}
\end{definition}
\begin{equation}
\begin{split}
PlanCost(q) \equiv dul.Quality(q) \ \ \land  \\ 
\exists p \ \ dul.Plan(p) \land hasCost(p,q). 
\end{split}
\label{eq:plan_cost}
\end{equation}
\begin{definition}
\textit{Plan Makespan is a Quality of a Plan that captures the expected time that would be required to execute the Plan.}
\end{definition}
\begin{equation}
\begin{split}
PlanMakespan(q) \equiv dul.Quality(q) \ \ \land \\ 
\exists p \ \ dul.Plan(p) \land hasMakespan(p,q).
\end{split}
\label{eq:plan_makespan}
\end{equation}
\begin{definition}
\textit{Plan Number Of Tasks is a Quality of a Plan that captures the number of tasks defined in the Plan.}
\end{definition}
\begin{equation}
\begin{split}
PlanNumberOfTasks(q) \equiv dul.Quality(q) \ \ \land \\ 
\exists p \ \ dul.Plan(p) \land hasNumberOfTasks(p,q).
\end{split}
\label{eq:plan_number_tasks}
\end{equation}

The formalization of `Plan Cost' (Eq.~\ref{eq:plan_cost}) reads as follows: a plan cost is a quality (q) that is the cost of at least one plan (p). The other two formalizations (Eq.~\ref{eq:plan_makespan} and Eq.~\ref{eq:plan_number_tasks}) would read in an analogous manner. The definitions include the notion of `executing a plan', used here as a primitive which means \textit{`following the sequence of tasks defined in the plan'}.  
The prefix \textit{`dul'} denotes that a term was re-used from DUL, while novel terms and relations have no prefix. The plan's properties are modeled as qualities (in DUL) that are related to a plan they qualify. New relations between the plans and the qualities were introduced: \textit{`has cost'}, \textit{`has makespan'} and \textit{`has number of tasks'}. Additionally, their inverse relations were also defined: \textit{`is cost of'}, \textit{`is makespan of'}, and \textit{`is number of tasks of'}. These two pairs of three new relations were defined as specializations of the DUL's relations \textit{`has quality'} and \textit{`is quality of'}, respectively.\looseness=-1

Of course, one might consider other qualities as relevant (e.g., the expected risk of human injury, the probability of failure/success, the workload percentage among collaborative agents, etc.). We argue that our approach is easy to extend to accommodate the specific details of other applications. This means that our approach can be extended with new qualities by just replicating the same structure that we propose. Consider the case of human-robot collaboration, a new plan’s quality might be the robot workload percentage (i.e. the percentage of the total plan’s tasks that the robot must do). This can also be defined as Quality and a new relation can be created e.g. hasRobotWorkload(p,w), used to relate a Plan (p) and the workload (w).

\subsubsection{How do the characteristics of different plans relate? (CQ2)} \label{sc:modeling_cq2}
The idea here is to be able to model knowledge such as: `the characteristic Xa of plan Pa is worse than the characteristic Xb of plan Pb'. OCRA does not provide a formal way to compare the properties of plans. Considering the previously formalized classes, the aimed relations should hold between qualities of plans (e.g. \textit{`PlanCost'}). Looking at the modeled qualities, one notices that all of them are numerical, hence, they could be related with comparative words such as: `higher', `lower', etc. However, it would be better to keep the new ontological model as re-usable as possible, for instance, using more generic notions (e.g. worse/better quality). Indeed, qualities between plans can be worse and better, but also equal or equivalent, thus, this should also be modeled. The aim of the new model is to provide the foundations for robots (and agents) to represent and reason about how alternative plans compare. Notions such as `better’ and `worse’ are indeed subjective, but that is exactly why it is relevant to work on their modeling. Hence, we propose a theory that copes with both numerical and categorical qualities.  

In total, three new intransitive relations are formalized: \textit{`is better quality than'}, \textit{`is worse quality than'}, \textit{`is equivalent quality to'}. The first two are inversely related, while the third one is symmetric. The three are defined as sub-relationships of the relation \textit{`associated with'} (from DUL). They hold between two qualities, thus, they can be used beyond the scope of this work (e.g. comparing the qualities of robots, drinks, etc.). The proposed ontology is designed to model the outcome of comparing two qualities. However, it falls outside the scope of the model to make any commitment regarding the comparison criteria. Sec.~\ref{sc:inference_rules_algorithm} proposes some general inference rules for this. Users of the ontology may use them or define their own.\looseness=-1

\subsubsection{How do different plans relate to each other? (CQ3)} \label{sc:modeling_cq3}
The OWL 2 DL version of OCRA defines the binary relationships \textit{`is better plan than'} and \textit{`is worse plan than'} which relate two plans stating that one of the plans is better or worse to achieve a goal. They were defined in the context of plan adaptations (i.e. events in which an agent decides to adapt an ongoing plan replacing it with a better option). Those two relations might be sufficient for the case of plan adaptations. Nevertheless, one might wonder what would happen in more general cases when two plans have equivalent properties. Neither of the plans would be better or worse than the other, thus, OCRA would fall short of modeling this. Furthermore, OCRA did not make any commitment about how an agent should compare plans and decide which one is better (see Sec.~\ref{sc:inference_rules_algorithm}).

Given the lack of a formalization in OCRA on this matter, this article extends its coverage by formalizing three new sub-relations, one per quality: \textit{`is cheaper plan than'}, \textit{`is faster plan than'}, \textit{`is shorter plan than'}; and their inverse relations \textit{`is more expensive plan than'}, \textit{`is slower plan than'}, \textit{`is longer plan than'}, respectively. All of them are defined as sub-properties of the relation \textit{`associated with'} (from DUL). Furthermore, it is also added the relation \textit{`is equivalent plan to'}, defined as disjoint with \textit{`is better plan than'} and \textit{`is worse plan than'}. Related to this new relation, other three relations are created as sub-properties of \textit{`associated with'}: \textit{`is plan with same cost as'}, \textit{`is plan with same makespan as'}, \textit{`is plan with same number of tasks as'}. Recall that all these relations hold between two plans, answering CQ3. 

\subsection{Formalization of the model in OWL 2 DL}
For practical use, the proposed ontological theory was formalized in OWL 2 DL. Hence, the axioms presented in this paper were translated to DL, in particular, to the SROIQ(D) fragment. Each of the axioms was translated in the target formalism with the exception of the value of plans' qualities. Note that quality is often used as a synonym for property but not in DUL, where `qualities are particulars and properties are universals'. In this regard, DUL considers that ‘qualities inhere in entities’~\cite{Gangemi_Guarino_Masolo_Oltramari_2003}. Every entity (including qualities) comes with its own exclusive qualities, which exist as long as the entity exists. DUL distinguishes between a quality (the cost of a plan) and its value or quale (a numerical data value). Hence, when saying that two plans have the same cost, their costs have the same quale, but still they are distinct qualities. This is convenient to model and answer CQ2, since OWL 2 DL cannot model relationships between two data values or quales (i.e. one cannot state that 5 is a better cost than 10). However, one can model the relation between the qualities (e.g. \textit{`cost A'} has better quality value than \textit{`cost B'}). Let's consider that a plan `p' has a cost `c' whose value is `10'. Hence, the knowledge would be modeled as:\looseness=-1  
\begin{center}
    $PlanCost(c) \land dul.Plan(p) \land hasCost(p,c) \land hasDataValue(c,10).$
\end{center} 

\begin{figure}[t!]
  \centering
  \includegraphics[width=1.0\linewidth]{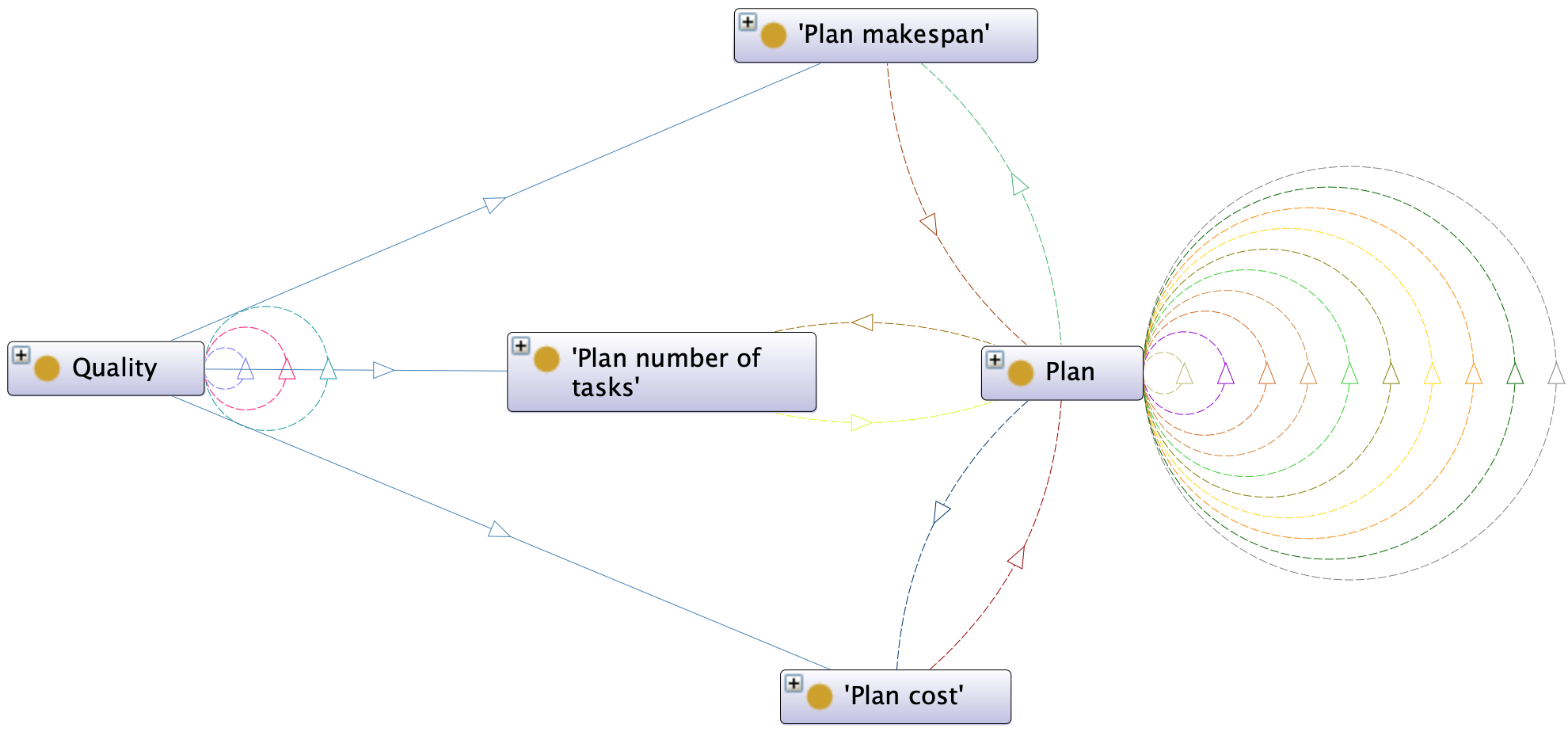}
  \caption{Graph visualization of the proposed ontological model. The defined plan's properties are subclasses of Quality, indicated by the only continuous arrows. The graph depicts the relations between plans and their properties (e.g., \textit{`has cost'}, \textit{`is cost of'}). It also shows multiple relations holding between plans (e.g. \textit{`is cheaper plan than'}), and three relations between qualities (e.g. \textit{`has better quality value than'}). }
  \label{fg:ontology_overview}
\end{figure}

For consistency, the label of the relations comparing two qualities: \textit{`is better quality than'}, \textit{`is worse quality than'}, \textit{`is equivalent quality to'}; were modified to \textit{`has better quality value than'}, \textit{`has worse quality value than'}, \textit{`has equivalent quality value than'}. Figure~\ref{fg:ontology_overview} shows an overview of the OWL 2 DL formalization of the ontology.   

\subsection{Modeling the tasks plans using DUL}
The sequence of tasks described in a plan is one of the useful aspects to compare plans. Therefore, modeling such knowledge would allow its use to narrate the differences between plans. The foundational ontology DUL already covers this knowledge, thus, there is no need for a new extension. As an example, let's imagine that there is a plan `p' that consists in executing three tasks in the following order, `t1', `t2', and `t3'. This knowledge might be represented as follows:\looseness=-1 

\begin{center}
    $ dul.Task(t1) \land dul.Task(t2) \land dul.Task(t3) \ \ \land $ \\ 
    $ dul.Plan(p) \land dul.definesTask(p,t1) \ \ \land $ \\
    $ dul.definesTask(p,t2) \land dul.definesTask(p,t3) \ \ \land $ \\
    $ dul.directlyFollows(t2,t1) \land dul.directlyFollows(t3,t2). $
\end{center}

\section{The theory at work}
\label{sc:theory_at_work}
The validation of an ontological model consists in creating instances of the concepts and showing their use. In this regard, the knowledge about instantiated plans (e.g., sequence of tasks, and qualities) would first be asserted to a knowledge base. Then, reasoning rules would be used to contrast the plans and infer which one is better. This section discusses the process of instantiating the model, and also introduces the reasoning rules used for contrasting the plans, and their implementation. Finally, it shows how the model is able to answer the competency questions, validating it. 

\subsection{Instantiating the ontology with plans}
\label{sc:instances}
The aim here is to demonstrate how to instantiate the model in a realistic scenario, providing as many resources as possible to potential users of the model. For this, the assertion of the knowledge about the plans was integrated with the planning system of the robot. Therefore, when a robot generates a plan by means of automated planning, it may also assert the knowledge about the plan, both its sequence and qualities. 
As a technical contribution, it was developed a novel knowledge-based framework that integrates existing robot planning tools and the use of the proposed ontology to model the comparison of plans (\textit{know-plan}). This implementation is publicly available on a Github repository,\footnote{\url{https://github.com/albertoOA/know_plan}} and illustrates how to instantiate the ontology with automatically generated plans. Planning is done using ROSPlan~\cite{cashmore2015rosplan}, a commonly used framework for planning in robotics. The ROSPlan framework provides a collection of tools for AI Planning for robots equipped with the Robot Operating System (ROS). Once a plan is found, the sequence of tasks and the plan's qualities are asserted to a knowledge base. As an example, a planning domain inspired by the scenario depicted in Fig.~\ref{fg:scenario} is used, where a robot delivers drinks at an office/apartment.\footnote{\url{https://github.com/albertoOA/know_plan/blob/main/pddl/domain/apartment_domain.pddl}} In order to have two plans to compare, the process is executed twice with different planning problems in which the type and location of the drink that the robot should bring change: a tea from the microwave or a cola from the fridge.\footnote{\url{https://github.com/albertoOA/know_plan/blob/main/pddl/problem/apartment_problem.pddl}} Since the tea and the coke are in different places, the planner finds plans with different costs, makespan and number of tasks (i.e., the cost of navigating to the coke is higher than to the tea, as it is further away). A summary of the asserted knowledge of the plans' qualities is presented in Tab.~\ref{tab:scenario_knowledge}.\looseness=-1

\begin{table}[t]
	\caption{Knowledge from the example of bringing drinks}
	\label{tab:scenario_knowledge}
    \centering
	\begin{tabular}{cccc}\toprule
		\textit{Plan}& \textit{Nº of Tasks}& \textit{Cost}& \textit{Makespan (s)}\\ \midrule
		`bringing tea'& 6& 27& 27\\
		`bringing cola'& 8& 59& 59\\
	\end{tabular}
\end{table}


\subsection{Reasoning for plan comparison}
\label{sc:inference_rules_algorithm}
The proposed ontological theory allows agents to represent the fact that a plan is better, worse or equally good than another one. However, none of the proposed axioms automates the comparison of plans and thus the inference of which plan is better. 

\subsubsection{Logical rules to infer the relation between plan's qualities} 
\label{sc:routine_compare_plans_cost}
Let's assume that there is a consistent instantiated ontology $\mathcal{O}$ that contains knowledge about the qualities of different plans ($P_a$, $P_b$) as a set of triples $\langle subject, relation, object \rangle$ (see Sec.~\ref{sc:instances}). The first step would be to compare the qualities' values and infer the relation between them (e.g. the cost of `bringing cola' has a worse value than the cost of `bringing tea'). Additionally, it can also be inferred the relation between the plans based on how the qualities relate (e.g. a plan is cheaper than another one). For instance, given the two plans ($P_a, P_b$), one can obtain their cost ($C_a, C_b$) and their cost' values ($V_a, V_b$) such that:  
\begin{center}
    $hasCost(P_a, C_a) \land hasCost(P_a, C_b) \land$ \\ 
    $dul.hasDataValue(C_a, V_a) \land dul.hasDataValue(C_b, V_b).$
\end{center}
Then, if the values are equal ($V_a = V_b$) two new triples would be added to the knowledge base indicating that both costs have an equivalent quality value and that both plans have the same cost:

\begin{center}
    $\mathcal{O} \leftarrow \mathcal{O} \cup \langle Ca, hasEquivalentQualityValueThan, Cb \rangle$; \\
    $\mathcal{O} \leftarrow \mathcal{O} \cup \langle Pa, isPlanWithSameCostAs, Pb \rangle.$
\end{center}
Similarly, when the values are different the asserted knowledge would refer to whether one of the costs/plans is worse/better than the other. Note that the cost's value is numerical, and it is usually assumed that the smaller it is, the better. Hence, when $V_a > V_b$:
\begin{center}
    $\mathcal{O} \leftarrow \mathcal{O} \cup \langle Ca, hasWorseQualityValueThan, Cb \rangle; $ \\
    $\mathcal{O} \leftarrow \mathcal{O} \cup \langle Pa, isMoreExpensivePlanThan, Pb \rangle. $
\end{center}
The final case would be when $V_a < V_b$, which would equally result in a triples' assertion of the inferred knowledge. As a whole, the complete described process becomes a logical rule to compare two plans' cost, which infers the relations between them and how this relation affects the connection between the plans. For the other qualities of plans formalized in the proposed model (makespan and number of tasks), analogous rules were defined. The criteria for the comparison were the same, since the rest of the formalized qualities are numerical and for all of them, the lower their value is, the better.

\subsubsection{Logical rule to infer which plan is better} Having inferred how the qualities and the plans relate, the next step is to infer which plan is better. Here, the criterion to decide if a plan is better than another one is satisfied when all the relations holding between their qualities indicate that the plan has better qualities. This is, if every quality of a plan ($Pa$) \textit{`has better quality value than'} the same type of quality of a second plan ($Pb$): 
\begin{center}
    $\forall Qa, Qb \ \ \exists Pa, Pb, R \ \ \langle Qb, rdf.type, dul.Quality \rangle \land \langle Qa, rdf.type, dul.Quality \rangle \land \langle Pa, R, Qa \rangle \land \langle Pb, R, Qb \rangle \land \langle Qa, hasBetterQualityValueThan, Qb \rangle$
\end{center}
The previous logical expression reads as follows: all pairs of qualities ($Qa$, $Qb$) related to a pair of plans ($Pa$, $Pb$) through the same property $R$, in which $Qa$ \textit{`has better quality value than'} $Qb$. The relation \textit{`rdf:type'}, which relates an individual with its class, is a standard relation from the Resource Description Framework (RDF)~\cite{McBride2004}. When the previous logical expression is true, then we can say that $Pa$ \textit{`is better plan than'} $Pb$, and the knowledge indicating it would be asserted:
\begin{center}
    $\mathcal{O} \leftarrow \mathcal{O} \cup \langle Pa, ocra.isBetterPlanThan, Pb \rangle. $
\end{center}
Hence, the chosen criterion to infer whether a plan is better than another plan is that all their properties shall be better. It is analogously defined for the cases of worse and equivalent plans, which as a whole would be the logical rule to infer, between two plans, which one is better. Note that in situations with conflicting or ambiguous properties, the logical rules will not assert new knowledge, thus it will not be specified whether one of the plans is equivalent to, better, or worse than the other one. For instance, if the cost of a plan has a better value than the cost of another plan, but in the rest of qualities the second plan is better, then none of the plans is inferred as better, neither equivalent nor worse. Note that the criterion is used here as an example, final users of the ontology might define different criteria. For instance, the effect of the different qualities might have a different weight (e.g. it may be more important to have a cheaper plan than a shorter one).


\subsection{Implementation of the inference rules}
\label{sc:inference_rules_implementation}
Inherited from \textit{know-cra}, this work uses Knowrob~\cite{tenorth2009knowrob,8460964}, a general framework for knowledge representation and reasoning for robots. The framework allows to read OWL-based ontologies and load them into a knowledge base that is built using Prolog~\cite{clocksin2012programming}. Since the knowledge base is accessible through a prolog-based interface, it is possible to use the logical reasoning power of Prolog to make inferences. Therefore, the inference rules introduced in Sec.~\ref{sc:inference_rules_algorithm} can be implemented in Prolog. This would integrate the decision-making process to compare plans into the knowledge base, augmenting the reasoning capabilities of the proposed ontological model. Note that the implementation of \textit{know-plan} also introduced some extra Prolog predicates to automate the call of the different rules. Specifically, it was implemented a predicate that runs all the rules for all the pairs of different plans stored in the knowledge base. The rules imply (binary) comparisons between pairs of qualities, and their complexity is linear with respect to the number of qualities, which would be added to the ontology (OWL 2 DL) complexity. 

\subsection{Answering the competency questions}
The example of a robot delivering drinks (see Sec.~\ref{sc:instances}) was used to showcase how to answer the competency questions. Note that the answers will contain the instantiated knowledge shown in Tab.~\ref{tab:scenario_knowledge}, plus inferred knowledge after applying the implemented rules. The queries are presented in prolog-like syntax (e.g. containing unbounded variables), since the knowledge base is written in Prolog.\looseness=-1

\subsubsection{Which are the characteristics of a plan? (CQ1)} \label{sc:answer_to_cq1} This competency question can be translated into a query in prolog-like syntax:

\begin{center}
    $triple('bringing \ \ tea', dul.hasQuality, Q), $ $triple(Q, dul.hasDataValue, V).$
\end{center}

If the query holds, i.e. the knowledge base contains the query triples, the answer contains an assignment of all the possible combinations of values of $Q$ and $V$ that make the query to be `true'. Some examples of answers would be: `cost of bringing tea' ($Q$) and `27' ($V$), or `number of tasks of bringing tea' ($Q$) and `6' ($V$).  

\subsubsection{How do the characteristics of different plans relate? (CQ2)} \label{sc:answer_to_cq2}

\begin{center}
    $triple('bringing \ \ tea', dul.hasQuality, Qa), $ $triple('bringing \ \ cola', dul.hasQuality, Qb), $ $triple(Qa, R, Qb).$
\end{center}

The answer to the query would contain an assignment of all the possible combinations of values of $Qa$, $R$ and $Qb$. For instance, `cost of bringing tea' ($Qa$), `has better quality value than' ($R$), and `cost of bringing cola' ($Qb$). Note that this can only be answered after some of the inference rules have been applied (see Sec.\ref{sc:routine_compare_plans_cost}).

\subsubsection{How do different plans relate to each other? (CQ3)} \label{sc:answer_to_cq3} 

\begin{center}
    $triple('bringing \ \ tea', R, 'bringing \ \ cola').$ 
\end{center}

The answer would contain all the assignments to $R$ that make the query to be `true' in the knowledge base. In total, $R$ could take four values: `is cheaper plan than', `is shorter plan than', `is faster plan than', and `is better plan than'. In order to answer this competency question, all the inference rules should have been applied.

\section{Contrastive narratives of plans}
In the literature, Miller~\cite{MILLER20191} discusses that the content of explanations is selected from all available knowledge and following some, often biased, criteria. One might argue that such a selection may be hard coded in our case, selecting a set of queries (e.g. the competency questions) to interrogate the knowledge base about how two plans compare. Then, the retrieved knowledge could be presented in a template (e.g. using a table or textual format). However, if the qualities change (e.g. new qualities are modeled), then the selection must be hard coded again. In this section, the objective is to discover a more flexible and advanced strategy to perform knowledge selection, retrieval, and compilation into contrastive explanatory narratives. 

\subsection{May explanatory narratives do the work?}
\label{sc:axon_limitations}

The ontological model proposed in this work augments the knowledge coverage in \textit{know-cra} to model the divergences between plans (see Sec.~\ref{sc:ontology}), and to automate the inference of whether those differences make one plan better than others (see Sec.~\ref{sc:theory_at_work}). Hence, one might think that using the new model together with the XONCRA methodology~\cite{10161359} would be enough to narrate what robots know about competing plans. In particular, given the knowledge about two plan instances and how they relate, XONCRA could produce a narrative about each of the plans using the AXON algorithm. The two narratives together would include the relevant knowledge for a robot to infer which plan is better, thus, humans could read the narratives and understand the inference. The differences between the two narratives could even be highlighted, as others have done when contrastively explaining the traces of two plans~\cite{10.1613/jair.1.12813}. However, such an approach would still require humans to extract their own conclusions by reading the complete narratives. Therefore, while being a potential baseline solution, the narratives generated by AXON do not seem to be optimized for the cases that concern this article, and a better approach might be developed.

\subsection{Beyond plain explanatory narratives}
\label{sc:beyond_xoncra}
Miller~\cite{MILLER20191} stated that explanations are \textit{contrastive}, \textit{selected}, and \textit{social}. Contrastive because they are sought in response to counterfactual cases that open questions such as: why a plan is better instead of others. Explanations are selected as they usually contain just part of the reasons, extracted by agents from a larger knowledge and based on specific criteria. Finally, explanations transfer knowledge in a conversational format, being part of a social interaction between agents. These three aspects of explanations set the basis to design a better algorithm, an alternative to AXON that:

\begin{itemize}
    \item constructs contrastive narratives instead of plain narratives;
    \item enhances the selection of knowledge, extracting only the differences between the compared plans; and 
    \item reduces the needed time to communicate a narrative, which might boost the (social) interaction. 
\end{itemize}

Naturally, the new algorithm shall preserve AXON's key advantages: being agnostic to the exact ontological objects and relationships, automating the selection and retrieval of relevant knowledge based on knowledge graph vicinity. Hence, the algorithm will remain applicable when new plan qualities are defined, and it will generalize beyond plan comparison (e.g. comparing two drinks, two tasks, etc.).

\subsection{Preliminary notation} 
\label{sc:preliminary_notation}
Let's assume countable pairwise disjoint sets $N_C$, $N_P$, and $N_I$ of class names, property names, and individuals, respectively. The standard relation \textit{`rdf:type'}, which relates an individual with its class, is abbreviated as \textit{`type'} and included in $N_P$. A knowledge graph $\mathcal{G}$ is a finite set of triples of the form $\langle s, p, o \rangle$ (subject, property, object), where $s \in N_I$, $p \in N_P$, $o \in N_I$ if $p \neq type$, and $o \in N_C$ otherwise. In this work, the knowledge base works as an episodic memory~\cite{8460964, 10161359}, thus it allows the assertion of triples with the time interval in which they hold. Hence, the stored knowledge can be seen as a time-indexed knowledge graph $\mathcal{G_T}$, which is a finite set of tuples of the form $\langle s, p, o, t_i, t_f \rangle$, where $t_i, t_f \in \mathbb{R} > 0$, and denote the time interval (initial and final time) in which the triple $\langle s, p, o \rangle$ holds. Note that non-asserted knowledge is considered unknown and never false, since knowledge graphs usually comply with the open-world assumption. For this, the Web Ontology Language 2 (OWL 2) was developed to allow the explicit negative assertion of properties: $\langle s, p, o \rangle$ is $false$. 
Hence, $\mathcal{G_T}$ may contain, e.g., that during an interval of time, $\langle t_i, t_f \rangle$, a task $k$ is not defined in a plan $p$: $\langle k, dul:isDefinedIn, p , t_i, t_f \rangle$ is $false$. In this work, querying the $\mathcal{G_T}$, we build what we called `contrastive narrative tuples' of a pair of instance plans, $\mathcal{T_P}$: $\langle s, p, o, t_i, t_f, sign \rangle$, where $sign$ indicates whether the time-indexed triple comes from a positive or negative assertion.\looseness=-1 

\subsection{ACXON - An algorithm for contrastive explanatory ontology-based narratives}
\label{sc:algorithm}
{\cAlgorithmName} is a novel theoretical contribution, an algorithm that retrieves knowledge about divergences between ontological entities (e.g. plans), for later construction of textual contrastive explanatory narratives (see Alg.~\ref{alg:acxon}). The algorithm takes as an input a time-indexed knowledge graph $\mathcal{G_T}$ (as described in Sec.~\ref{sc:preliminary_notation}), the ontological class (or classes) of the pair of instances to narrate, the temporal locality (time interval of interest), and the level of specificity. Our focus is on contrastive narratives about \textit{Plans}, but {\cAlgorithmName} is general enough to work with other OWL 2 DL ontologies and classes. For instance, it might contrastively narrate the capabilities of a pair of agents (e.g. one can move for longer periods), or how two drinks are different to each other (e.g. one is healthier, tastier, etc.). Based on the level of specificity, there are three types of contrastive narratives. In this article, specificity refers to the amount of detail to be used during the narrative construction, i.e., the number of knowledge tuples.\looseness=-1  

\begin{algorithm}[tb!]
 \KwIn{
 Time-indexed knowledge graph ($\mathcal{G_T}$), pairs to narrate ($\mathcal{P}$), temporal locality ($L_i, L_f$), specificity ($S$)}
 \KwOut{Contrastive Narratives ($\mathcal{E}$)}
 $\mathcal{E}$ $\xleftarrow{}$ $\emptyset$\\ 
 $\mathcal{I_{P_T}}$ $\xleftarrow{}$ RetrieveInstantiatedPairsWithTime($\mathcal{G_T}$, $\mathcal{P}, L_i, L_f$)\\
 \ForEach{$\langle \langle e_a, t_{i_a}, t_{f_a} \rangle , \langle e_b, t_{i_b}, t_{f_b} \rangle \rangle \in \mathcal{I_{P_T}}$}{
    $\mathcal{I_P} \xleftarrow{} \langle \langle e_a, t_{i_a}, t_{f_a} \rangle , \langle e_b, t_{i_b}, t_{f_b} \rangle \rangle$ \\
    $\mathcal{T_P}$ $\xleftarrow{}$ RetrieveNarrativeTuples($\mathcal{G_T}, \mathcal{I_P}, S$)\\
    $\mathcal{D_P}$ $\xleftarrow{}$ ExtractDivergentNarrativeTuples($\mathcal{T_P}$)\\
    $\mathcal{E_P}$ $\xleftarrow{}$ ConstructContrastiveNarrative($\mathcal{D_P}$)\\
    $\mathcal{E}$ $\xleftarrow{}$ $\mathcal{E} \cup \mathcal{E_P}$\\
 }
 
 \caption{{\cAlgorithmName}}
 \label{alg:acxon}
\end{algorithm}

{\cAlgorithmName} first retrieves a set $\mathcal{I_{P_T}}$ of sets comprising the instantiated pairs of the provided pair of classes $\mathcal{P}$, which exist, at least partially, within the temporal locality $\langle L_i, L_f \rangle$ (line 2). Each $\mathcal{I_P}$ is a set of tuples $\langle e, t_i, t_f \rangle$, containing the two pair's instances $e$ with their time interval $(t_i, t_f)$ (line 4). Second, for each of the instantiated pairs $\langle \langle e_a, t_{i_a}, t_{f_a} \rangle , \langle e_b, t_{i_b}, t_{f_b} \rangle \rangle \in \mathcal{I_{P_T}}$, a set of knowledge tuples $\mathcal{T_P}$ is retrieved according to the specificity level (line 5). Third, from the initial narrative tuples $\mathcal{T_P}$, it is selected the sub-set containing only divergent knowledge between the pair of instances $\mathcal{D_P}$ (line 6). Fourth, for each instantiated pair a contrastive explanation $\mathcal{E_P}$ is built using their relative tuples (line 7). Finally, the new explanation is added to the set of explanations $\mathcal{E}$ (line 8). The implemented algorithm and examples of its use are available online.\footnote{\url{www.iri.upc.edu/groups/perception/\#ontology-based-explainable-robots}}\looseness=-1

\subsubsection{Retrieve instantiated pair with time routine} 
\label{sc:routine1}
With a time-indexed knowledge graph $\mathcal{G_T}$, a pair of ontological existing classes or a set of them, $\mathcal{P} \subset N_C$, and the temporal locality $\langle L_i, L_f\rangle$, this routine retrieves a set of sets $\mathcal{I_{P_T}}$ comprising all the instantiated pairs of the given pair of classes, $\mathcal{I_P}$, which contain the two time-indexed instances $\langle \langle e_a, t_{i_a}, t_{f_a} \rangle , \langle e_b, t_{i_b}, t_{f_b} \rangle \rangle$ of each pair such that their time intervals exist within (intersect) the temporal locality:
\begin{center}
    $\forall \langle \langle e_a, t_{i_a}, t_{f_a} \rangle , \langle e_b, t_{i_b}, t_{f_b} \rangle \rangle \in \mathcal{I_{P_T}} \rightarrow \exists \langle c_a, c_b \rangle \in \mathcal{P} \land $ 
    $ \langle e_a, type, c_a, t_{i_a}, t_{f_a}, sign_a \rangle \in \mathcal{G_T} \land (\langle t_{i_a}, t_{f_a} \rangle \cap \langle L_i, L_f \rangle) \land $
    $ \langle e_b, type, c_b, t_{i_b}, t_{f_b}, sign_b \rangle \in \mathcal{G_T} \land (\langle t_{i_b}, t_{f_b} \rangle \cap \langle L_i, L_f \rangle) $ .
\end{center}
Examples of time-indexed instances of plans to narrate may be:

\vspace{2pt}
\noindent\fbox{%
    \parbox{0.96\columnwidth}{%
        {\footnotesize
        $\langle \langle bringing \ \ water, 10.0, 150.0\rangle$, $\langle bringing \ \ juice, 0.0, 150.0\rangle \rangle$; \\
        $\langle \langle bringing \ \ tea, $ \_ $, Inf\rangle$, $\langle bringing \ \ cola, $ \_ $, Inf\rangle \rangle$.
        }
    }%
}
\vspace{2pt}

Note that the time interval is not always numerical, see the second example. This happens when a triple has held true in the knowledge base since an undetermined instant of time (`\_'), and it will remain true as long as the knowledge base stays active (`$Inf$').

\subsubsection{Retrieve narrative tuples routine} 
\label{sc:routine2}
For each instantiated pair to narrate $\mathcal{I_P}$, with the $\mathcal{G_T}$, and the level of specificity $S$, the routine would retrieve all the tuples about the pair, $\langle s, p, o, t_i, t_f, sign \rangle$, that are relevant to build the contrastive narrative. 
The first level of specificity retrieves tuples comprising all the relations $p$ holding between the two instances of a pair: $\exists p  \in N_P \land (\langle e_a, p, e_b, t_i, t_f, sign \rangle \in \mathcal{G_T} \lor \langle e_b, p, e_a, t_i, t_f, sign \rangle \in \mathcal{G_T})$. 
In the second level, the routine adds all the tuples in which at least one of the instances $\langle e_a, e_b \rangle$ is related through a property $p$ to an object $o$: $\exists p \in N_P \land (\langle e_a, p, o, t_i, t_f, sign \rangle \in \mathcal{G_T} \lor \langle e_b, p, o, t_i, t_f, sign \rangle \in \mathcal{G_T})$. Following a similar logic to the first level, the second level also retrieves tuples that relate the different objects $o$. When the instances $\langle e_a, e_b \rangle$ are related to two objects $\langle o_a, o_b \rangle$ through the same property $p$, the tuples relating those two objects $\langle o_a, q, o_b, t_i, t_f, sign \rangle$ are added: $\exists p, q \in N_P \land \langle o_a, q, o_b, t_i, t_f, sign \rangle \land \langle e_a, p, o_a, t_{i_a}, t_{f_a}, sign_a \rangle \in \mathcal{G_T} \land \langle e_b, p, o_b, t_{i_b}, t_{f_b}, sign_b \rangle \in \mathcal{G_T}$. These are horizontal links in Fig.~\ref{fg:graph_specificity}. Then, in the third level the routine adds all the tuples in which the objects $o$ from the second level are related to other objects $o_x$:  $\langle o, p_x, o_x, t_{ix}, t_{fx}, sign_x \rangle \in \mathcal{G_T} \land \langle t_i, t_f \rangle \cap \langle t_{ix}, t_{fx} \rangle$. Robots' experiences may be tied to a time frame, thus, the search was restricted to tuples whose time interval $\langle t_{ix}, t_{fx}\rangle$ intersected the time interval of the pair's instances $\langle t_{i_a}, t_{f_a} \rangle$ and $\langle t_{i_b}, t_{f_b} \rangle$. This prevented the routine from retrieving tuples with irrelevant knowledge about the pair of instances to narrate $\langle e_a, e_b \rangle$. The third level finishes collecting the tuples relating the different objects $o_x$ between each other, similarly to how it is done in the second level: $\exists p_x, q_x \in N_P \land \langle o_{x_a}, q_x, o_{x_b}, t_i, t_f, sign \rangle \land \langle o_a, p_x, o_{x_a}, t_{i_a}, t_{f_a}, sign_{x_a} \rangle \in \mathcal{G_T} \land \langle o_b, p_x, o_{x_b}, t_{i_b}, t_{f_b}, sign_{x_b} \rangle \in \mathcal{G_T}$. For any of the levels, when the narrative's set of tuples $\mathcal{T_P}$ already contains a tuple or its inverse, the tuple is not added. Furthermore, the tuples are retrieved incrementally from the first to the third level. Hence, when the specificity level is three, the returned tuples also contain those from the first and second levels. This would equate to moving deeper in the knowledge graph representing the instanced pair (see Fig. \ref{fg:graph_specificity}). 
Utilizing the ongoing example of bringing a drink (see Fig.~\ref{fg:scenario}), some instances of the retrieved narrative tuples $\mathcal{T_P}$ of an instantiated pair are:\looseness=-1

\vspace{2pt}
\noindent\fbox{%
    \parbox{0.96\columnwidth}{%
        {\footnotesize
        $\boldsymbol{\mathcal{T_P}_{1}}$ $\langle$`bringing tea', isBetterPlanThan, `bringing cola', \_, Inf, positive$\rangle$, \\
        $\boldsymbol{\mathcal{T_P}_{2}}$ $\langle$`bringing tea', definesTask, `T2-grasp object', \_, Inf, positive$\rangle$, \\
        $\boldsymbol{\mathcal{T_P}_{3}}$ $\langle$`bringing tea', definesTask, `task 0 - find person', \_, Inf, positive$\rangle$, \\
        $\boldsymbol{\mathcal{T_P}_{4}}$ $\langle$`bringing cola', definesTask, `task 0 - find person', \_, Inf, positive$\rangle$, \\
        $\boldsymbol{\mathcal{T_P}_{5}}$ $\langle$`T3-go to waypoint', directlyPrecedes, `T5-give object', \_, Inf, positive$\rangle$, \\
        $\boldsymbol{\mathcal{T_P}_{6}}$ $\langle$`T7-give object', isTaskDefinedIn, `bringing cola', \_, Inf, positive$\rangle$, \\
        $\boldsymbol{\mathcal{T_P}_{7}}$ $\langle$`bringing tea', definesTask, `T3-got to waypoint', \_, Inf, positive$\rangle$, \\
        $\boldsymbol{\mathcal{T_P}_{8}}$ $\langle$`bringing tea', definesTask, `T5-give object', \_, Inf, positive$\rangle$, \\
        $\boldsymbol{\mathcal{T_P}_{9}}$ $\langle$`T3-go to waypoint', directlyFollows, `T2-grasp object', \_, Inf, positive$\rangle$, \\
        $\boldsymbol{\mathcal{T_P}_{10}}$ $\langle$`bringing tea', isCheaperPlanThan, `bringing cola', \_, Inf, positive$\rangle$, \\
        $\boldsymbol{\mathcal{T_P}_{11}}$ $\langle$`bringing cola', hasCost, `cola cost', \_, Inf, positive$\rangle$, \\
        $\boldsymbol{\mathcal{T_P}_{12}}$ $\langle$`bringing tea', hasCost, `tea cost', \_, Inf, positive$\rangle$, \\
        $\boldsymbol{\mathcal{T_P}_{13}}$ $\langle$`cola cost', hasDataValue, `59', \_, Inf, positive$\rangle$, \\
        $\boldsymbol{\mathcal{T_P}_{14}}$ $\langle$`tea cost', hasDataValue, `27', \_, Inf, positive$\rangle$, \\
        $\boldsymbol{\mathcal{T_P}_{15}}$ $\langle$`cola cost', hasWorseQualityValueThan, `tea cost', \_, Inf, positive$\rangle$.
        }
    }%
}
\vspace{2pt}

\subsubsection{Extract divergent narrative tuples} 
\label{sc:routine3}
\begin{figure}[t!]
  \centering
  \includegraphics[width=1.0\linewidth]{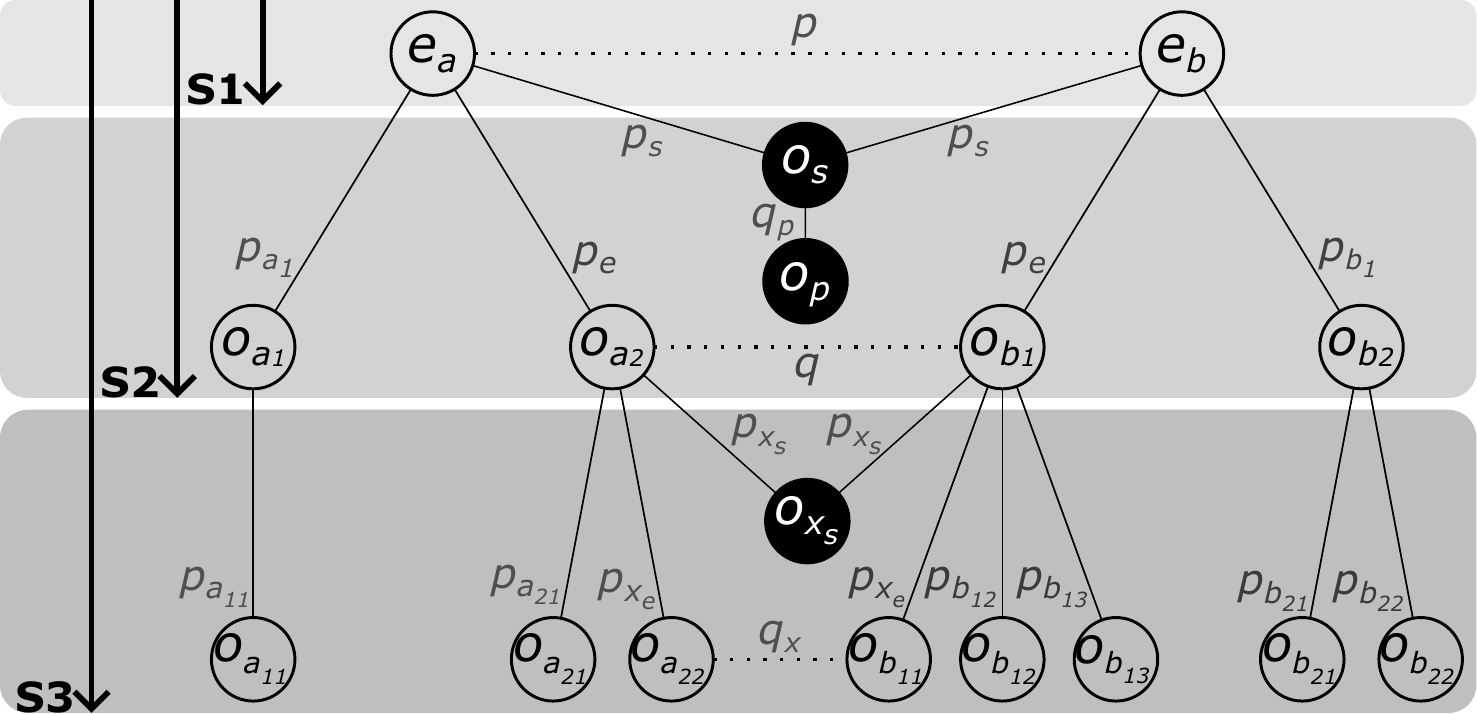}
  \caption{Graphical representation of the different levels of specificity (S1, S2, S3) and their respective depth in the knowledge graph. Nodes in black correspond to initially retrieved tuples that are later pruned because they are non-divergent. S1 contains direct relations between the ontology instances to be compared (e.g. Plan A is cheaper than Plan B). S2 compares how the target instances relate to other objects, and the relations between those objects (e.g. Plan A has a specific cost (Cost A) while Plan B has a different cost (Cost B), or Cost A has better quality value than Cost B). S3 goes deeper in the knowledge graph, comparing the relationships of the objects of the previous level with other ontological objects (e.g. Cost A has data value 27).}
  \label{fg:graph_specificity}
\end{figure}

From the initially selected narrative tuples $\mathcal{T_P}$, the routine would just retrieve the set of knowledge tuples $\mathcal{D_P}$ that capture divergences between the two pair's instances. The routine identifies the non-divergent tuples that exist in $\mathcal{T_P}$ and prunes them. A pair of tuples $\langle \langle s_1, p_1, o_1, t_{i_1}, t_{f_1}, sign_1 \rangle$, $\langle s_2, p_2, o_2, t_{i_2}, t_{f_2}, sign_2 \rangle \rangle \in \mathcal{T_P}$ will be non-divergent when: $(s_1 \neq s_2) \land (p_1 = p_2) \land (o_1 = o_2) \land (t_{i_1} = t_{i_2}) \land (t_{f_1} = t_{f_2}) \land (sign_1 = sign_2)$. Note that the routine prunes the whole branch of a non-divergent tuple, which includes tuples in which the shared object $(o_1 = o_2 = o_s)$ acts as the subject: $\langle o_s, q_p, o_p, t_{i}, t_{f}, sign \rangle$. The process will apply to tuples extracted at any of the specificity levels, as it is depicted in Fig.~\ref{fg:graph_specificity}. In the example, after applying this routine, the tuples $\mathcal{T_P}_{3}$ and $\mathcal{T_P}_{4}$ would be pruned, and the set of remaining tuples would be:\looseness=-1

\vspace{2pt}
\noindent\fbox{%
    \parbox{0.96\columnwidth}{%
        {\footnotesize
        $\boldsymbol{\mathcal{D_P}_{1}}$ $\langle$`bringing tea', isBetterPlanThan, `bringing cola', \_, Inf, positive$\rangle$, \\
        $\boldsymbol{\mathcal{D_P}_{2}}$ $\langle$`bringing tea', definesTask, `T2-grasp object', \_, Inf, positive$\rangle$, \\
        $\boldsymbol{\mathcal{D_P}_{3}}$ $\langle$`T3-go to waypoint', directlyPrecedes, `T5-give object', \_, Inf, positive$\rangle$, \\
        $\boldsymbol{\mathcal{D_P}_{4}}$ $\langle$`T7-give object', isTaskDefinedIn, `bringing cola', \_, Inf, positive$\rangle$, \\
        $\boldsymbol{\mathcal{D_P}_{5}}$ $\langle$`bringing tea', definesTask, `T3-got to waypoint', \_, Inf, positive$\rangle$, \\
        $\boldsymbol{\mathcal{D_P}_{6}}$ $\langle$`bringing tea', definesTask, `T5-give object', \_, Inf, positive$\rangle$, \\
        $\boldsymbol{\mathcal{D_P}_{7}}$ $\langle$`T3-go to waypoint', directlyFollows, `T2-grasp object', \_, Inf, positive$\rangle$, \\
        $\boldsymbol{\mathcal{D_P}_{8}}$ $\langle$`bringing tea', isCheaperPlanThan, `bringing cola', \_, Inf, positive$\rangle$, \\
        $\boldsymbol{\mathcal{D_P}_{9}}$ $\langle$`bringing cola', hasCost, `cola cost', \_, Inf, positive$\rangle$, \\
        $\boldsymbol{\mathcal{D_P}_{10}}$ $\langle$`bringing tea', hasCost, `tea cost', \_, Inf, positive$\rangle$, \\
        $\boldsymbol{\mathcal{D_P}_{11}}$ $\langle$`cola cost', hasDataValue, `59', \_, Inf, positive$\rangle$, \\
        $\boldsymbol{\mathcal{D_P}_{12}}$ $\langle$`tea cost', hasDataValue, `27', \_, Inf, positive$\rangle$, \\
        $\boldsymbol{\mathcal{D_P}_{13}}$ $\langle$`cola cost', hasWorseQualityValueThan, `tea cost', \_, Inf, positive$\rangle$.
        }
    }%
}
\vspace{2pt}

\subsubsection{Construct contrastive narrative routine} 
\label{sc:routine4}
Considering the divergent tuples $\mathcal{D_P}$ of a pair of instances $\langle e_a, e_b \rangle$ to narrate, the routine builds the contrastive explanatory narrative applying a set of rules: \textbf{casting}, \textbf{clustering}, \textbf{ordering}, and \textbf{grouping}. These rules describe aggregations commonly used by humans in natural language~\cite{10.1007/3-540-60800-1_25}.\looseness=-1

\textbf{Casting} consists in homogenizing the ontological properties appearing in the tuples. First, ensuring that for all the tuples $\mathcal{D_P}$ that concern any of the pair's instances $\langle e_a, e_b \rangle$ the instances are the tuple's subject. In the ongoing example: $\mathcal{D_P}_{1}$, $\mathcal{D_P}_{2}$, $\mathcal{D_P}_{4}$, $\mathcal{D_P}_{5}$, $\mathcal{D_P}_{6}$, $\mathcal{D_P}_{8}$, $\mathcal{D_P}_{9}$, and $\mathcal{D_P}_{10}$. Hence, when  $\mathcal{D_P}$ contains a tuple in which any of the instances $e$ acts as the object, $\langle s, p, e, t_i, t_f, sign \rangle \in \mathcal{D_P}$, the casting rule reverses the tuple to: $\langle e, p^{-1}, s, t_i, t_f, sign \rangle$, where $p^{-1}$ states for the inverse property of $p$. In the list of tuples from before, the tuple $\mathcal{D_P}_{4}$ that contains the property \textit{`isTaskDefinedIn'} would be reversed using \textit{`definesTask'}. Then, all the tuples concerning any of the two instances, both reversed tuples and those that did not need to be inverted, are added to a new set $\mathcal{D_P}_{Cast}$ of cast tuples. 
Casting has a second step that involves the tuples not concerning the pair's instances ($\mathcal{D_P}_{3}$, $\mathcal{D_P}_{7}$, $\mathcal{D_P}_{11}$, $\mathcal{D_P}_{12}$, and $\mathcal{D_P}_{13}$). Guaranteeing that the properties of the tuples to add are consistent with those that already exist in the cast tuples. Hence, if it is not the case, the tuple is reversed before adding it to $\mathcal{D_P}_{Cast}$. In the example, $\mathcal{D_P}_{3}$ is added to $\mathcal{D_P}_{Cast}$ (following the order) thus, $\mathcal{D_P}_{7}$ needs to be inverted before added. $\mathcal{D_P}_{11}$, $\mathcal{D_P}_{12}$, $\mathcal{D_P}_{13}$ are just added.\looseness=-1

Next, this routine applies \textbf{clustering}, which structures the tuples in a way that each cluster will later be used to form a single sentence within the whole narrative. The narratives shall contrast the knowledge relating the pair's instances, revealing the divergences between them. Therefore, an effective strategy for cluster generation is to utilize the structure of the knowledge graph formed by the retrieved tuples (see Fig.~\ref{fg:graph_specificity}). First, a cluster is created with the tuples $\langle e_a, p, e_b, t_i, t_f, sign \rangle$ that relate the pair's instances $\langle e_a, e_b\rangle$, in the example, $\mathcal{D_P}_1$ and $\mathcal{D_P}_8$. Second, the routine clusters the remaining tuples by property $p$ in three different steps named: \textit{direct}, \textit{indirect}, \textit{unrelated}. The tuples \textit{directly} related through $p$ to the instances to compare: $\langle e_a, p, o_a, t_i, t_f, sign \rangle$, $\langle e_b, p, o_b, t_i, t_f, sign \rangle$, are clustered together with the tuples relating their objects: $\langle o_a, q, e_b, t_i, t_f, sign \rangle$. Note that when only one of the instances $\langle e_a, e_b\rangle$ is related to an object through $p$, e.g., $\langle e_a, p, o_a, t_i, t_f, sign \rangle \in \mathcal{D_P} \land \langle e_b, p, o_b, t_i, t_f, sign \rangle \notin \mathcal{D_P}$; the knowledge will be clustered later at the \textit{unrelated} step. In the example, $\mathcal{D_P}_{9}$, $\mathcal{D_P}_{10}$, $\mathcal{D_P}_{13}$ form a cluster, and $\mathcal{D_P}_2$, $\mathcal{D_P}_5$, $\mathcal{D_P}_6$ and the reversed $\mathcal{D_P}_4$ another one. New clusters are created for the tuples \textit{indirectly} related to the instances to compare, i.e. those related to the objects $\langle o_a, o_b\rangle$ of the previous step: $\langle o_a, p_x, o_{x_a}, t_{i_a}, t_{f_a}, sign_{x_a} \rangle$, $\langle o_b, p_x, o_{x_b}, t_{i_b}, t_{f_b}, sign_{x_b} \rangle$. As before, those clusters include the tuples holding between their objects $\langle o_{x_a}, o_{x_b} \rangle$. Recall that the tuples are only clustered if the objects $\langle o_a, o_b\rangle$ are each related to one of the main instances to compare. In the example, $\mathcal{D_P}_{11}$ and $\mathcal{D_P}_{12}$ form a cluster. Finally, (\textit{unrelated}) clusters are created with the remaining tuples sharing the same property $p$, in the example, $\mathcal{D_P}_{3}$, and the inverted $\mathcal{D_P}_{7}$ are clustered.

Subsequently, the clustered tuples are \textbf{ordered} externally (between clusters) and internally (between tuples). When externally ordering, the set of clusters is ordered according to the sequence followed during the clustering: first the cluster relating the pair's instances followed by the clusters obtained in the \textit{direct}, \textit{indirect}, and \textit{unrelated} steps. Then, within the clusters from a single step, the clusters are ordered from more knowledge (more tuples) to less. The internal ordering just assures that the tuples with the property $p = type$ are at the front of each of the clusters. In the example, after applying all these rules the set of tuples would now be:\looseness=-1

\vspace{2pt}
\noindent\fbox{%
    \parbox{0.96\columnwidth}{%
        {\footnotesize
        $\boldsymbol{\mathcal{D_P}_{1}}$ $\langle$`bringing tea', isBetterPlanThan, `bringing cola', \_, Inf, positive$\rangle$, \\
        $\boldsymbol{\mathcal{D_P}_{2}}$ $\langle$`bringing tea', isCheaperPlanThan, `bringing cola', \_, Inf, positive$\rangle$, \\ 
        
        $\boldsymbol{\mathcal{D_P}_{3}}$ $\langle$`bringing tea', definesTask, `T2-grasp object', \_, Inf, positive$\rangle$, \\
        $\boldsymbol{\mathcal{D_P}_{4}}$ $\langle$`bringing tea', definesTask, `T3-got to waypoint', \_, Inf, positive$\rangle$, \\
        $\boldsymbol{\mathcal{D_P}_{5}}$ $\langle$`bringing tea', definesTask, `T5-give object', \_, Inf, positive$\rangle$, \\
        $\boldsymbol{\mathcal{D_P}_{6}}$ $\langle$`bringing cola', definesTask, `T7-give object', \_, Inf, positive$\rangle$, \\
        
        $\boldsymbol{\mathcal{D_P}_{7}}$ $\langle$`bringing cola', hasCost, `cola cost', \_, Inf, positive$\rangle$, \\
        $\boldsymbol{\mathcal{D_P}_{8}}$ $\langle$`bringing tea', hasCost, `tea cost', \_, Inf, positive$\rangle$, \\
        $\boldsymbol{\mathcal{D_P}_{9}}$ $\langle$`cola cost', hasWorseQualityValueThan, `tea cost', \_, Inf, positive$\rangle$, \\
        
        $\boldsymbol{\mathcal{D_P}_{10}}$ $\langle$`cola cost', hasDataValue, `59', \_, Inf, positive$\rangle$, \\
        $\boldsymbol{\mathcal{D_P}_{11}}$ $\langle$`tea cost', hasDataValue, `27', \_, Inf, positive$\rangle$, \\
        
        $\boldsymbol{\mathcal{D_P}_{12}}$ $\langle$`T3-go to waypoint', directlyPrecedes, `T5-give object', \_, Inf, pos.$\rangle$, \\
        $\boldsymbol{\mathcal{D_P}_{13}}$ $\langle$`T2-grasp object', directlyPrecedes, `T3-go to waypoint', \_, Inf, pos.$\rangle$.
        }
    }%
}
\vspace{2pt}

Finally, the tuples of each cluster pass through several \textbf{grouping} steps, obtaining the sentences of the final textual contrastive narrative $\mathcal{E_P}$. First, object grouping, the tuples sharing subject, property, interval, and sign are united. Thus, given $\langle s, p, o_a, t_i, t_f, sign \rangle$ and $\langle s, p, o_b, t_i, t_f, sign \rangle$, at this step the routine unites them to: $\langle s, p, o_a \ and \  o_b, t_i, t_f, sign \rangle$. In the example, $\mathcal{D_P}_3$, $\mathcal{D_P}_4$, and $\mathcal{D_P}_5$ are grouped into: $\langle$`bringing tea', definesTask, `T2 - grasp object' and `T3 - got to waypoint' and `T5 - give object', \_, Inf, positive$\rangle$. The second step consists in grouping tuples by predicate (property), thus tuples sharing subject, object, interval, and sign are joined. Considering two tuples: $\langle s, p_a, o, t_i, t_f, sign \rangle$ and $\langle s, p_b, o, t_i, t_f, sign \rangle$, the routine unites them to: $\langle s, p_a \ and \ p_b, o, t_i, t_f, sign \rangle$. In the example, $\mathcal{D_P}_1$ and $\mathcal{D_P}_2$ would be grouped. The final grouping stage generates textual contrastive sentences for the knowledge stored in each of the clusters of tuples by translating the tuples into text and connecting them. For instance, a cluster may contain the tuples $\langle s, p, o, t_i, t_f, sign \rangle$ and $\langle o, q, o_x, t_{i_x}, t_{f_x}, sign \rangle$. Hence, the knowledge is connected as a subordinate sentence using the pronoun `which'. In the example, this happens with $\mathcal{D_P}_7$ and $\mathcal{D_P}_{9}$. Note that when the subordinate is introduced in tuples that have gone through object grouping, it is also added the phrase `and also' for readability purposes. The explanations are contrastive, thus the conjunction `while' is added to emphasize the divergent (contrastive) knowledge: e.g. $\langle e_a, p, o_a, t_{i_a}, t_{f_a}, sign_a \rangle$ and $\langle e_b, p, o_b, t_{i_b}, t_{f_b}, sign_b \rangle$. In the example, `while' is used to compare the knowledge from the grouped $\mathcal{D_P}_7$ and $\mathcal{D_P}_{9}$, and $\mathcal{D_P}_8$; and $\mathcal{D_P}_6$ and the grouped $\mathcal{D_P}_3$, $\mathcal{D_P}_4$, $\mathcal{D_P}_5$. The the same applies for \textit{indirectly} related clusters such as the one including tuples $\mathcal{D_P}_{10}$ and $\mathcal{D_P}_{11}$. The tuples from \textit{unrelated} clusters, e.g. $\mathcal{D_P}_{12}$ and $\mathcal{D_P}_{13}$, are connected using `and'. The propositions `from' and `to' are also added to introduce the tuples' time intervals, but only if they are different to the interval of the pair's instances. Indeed, if the interval is undetermined (\_, Inf), it is obviated. The names of ontological entities (instances, classes and properties) are kept, only some properties are slightly changed to more understandable terms (e.g. using `includes task' instead of `definesTask', or `has a higher value than' instead of `hasWorseQualityValueThan'). The final narrative for the ongoing example would be:\looseness=-1

\vspace{2pt}
\noindent\fbox{%
    \parbox{0.96\columnwidth}{%
        {\footnotesize
        `bringing tea' is better plan than and is cheaper plan than `bringing cola'. 
        `bringing tea' includes task `T2-grasp object' and `T3-got to waypoint' and `T5-give object', while `bringing cola' includes task `T7-give object'. `bringing cola' has cost `cola cost', which has a higher value than `tea cost'; while `bringing tea' has cost `tea cost'. 
        `cola cost' has value `59', while `tea cost' has value `27'. 
        `T3-go to waypoint' directly precedes `T5-give object', and `T2-grasp object' directly precedes `T3-go to waypoint'.
        }
    }%
}
\vspace{2pt}

\section{Evaluating explanatory narratives}
\label{sc:evaluation_general_ref}
To evaluate the quality of the narratives generated by {\cAlgorithmName}, the AXON~\cite{10161359} algorithm was used as a baseline. Specifically, both algorithms were used to narrate the knowledge about contrasting plans. Following the ideas discussed in Sec.~\ref{sc:axon_limitations}, these two algorithms can be compared by using AXON twice (i.e. to narrate each of the plans independently). By construction, {\cAlgorithmName} is expected to reduce the amount of knowledge used in the explanations, and also to ensure shorter explanation communication times. 

\subsection{Evaluation procedure and setup}
\label{sc:evaluation_procedure}
The evaluation was done using a set of temporal planning PDDL domains from recent international planning competitions (IPC)~\cite{Fox2003The3I}. The set included the IPC'02 Rovers domain~\cite{Fox2003The3I} (10 instantiated problems), the IPC'08 Crew planning domain~\cite{4839709} (15 problems), and the IPC'14 Match cellar domain~\cite{halsey2004crikey} (20 problems). First, given a domain and two problems, we run a planner with both problems to obtain two plans for which their respective knowledge (sequence and qualities) is instantiated as described in Sec.~\ref{sc:instances}. Then, the inference rules are applied, performing the comparison of the two plans and asserting the inferred knowledge (e.g. which plan is better). Finally, both algorithms are used to extract the knowledge from the active knowledge base and construct the explanations. The algorithms have the same inputs: a time-indexed graph (the active knowledge base), a time interval (\_, Inf) and the level of specificity (the three levels were used). A set of metrics discussed in Sec.~\ref{sc:metrics} are computed for each of the generated narratives. Note that for each of the planning domains, ten pairs of instances were randomly selected without replacement (thirty in total). The software for the test was run on a desktop PC with an Intel Core i7-8700K CPU (12x 3.70 GHz), 16 GB DDR4 RAM, and an NVIDIA GeForce GT 710/PCIe/SSE2 GPU. 

\subsection{Metrics for explanation evaluation}
\label{sc:metrics}
To evaluate the explanations we have selected a set of offline objective evaluation metrics, aligned with the existing literature. The metrics aim to evaluate two of the main features of explanations: the selection of content (number of attributes), and the social aspect (communication time and readability). \\
\textbf{Number of attributes.} The metric is commonly used to evaluate explainable models. Especially when evaluating the explainability of black box models (e.g. machine learning (ML) models)~\cite{10.5555/3463952.3463962}. Nevertheless, this metric has also been used to evaluate non-ML explanatory systems~\cite{10.5555/3535850.3535909}. In this work, the number of attributes is equal to the number of tuples $\mathcal{D_P}$ used to construct the narratives.\looseness=-1 \\
\textbf{Communication time.} Explanations are social, thus a good quality index is to measure how much time would require an agent to communicate them. In this work, the communication time is computed as a combination of the \textit{construction time} $C_T$ and the actual \textit{interaction time} $I_T$. For the interaction, two channels are considered: auditory (${I_T}_A$) and visual (${I_T}_V$). For retaining information, people are comfortable with a speaking pace of 150-160 words per minute (wpm), while the pace for silent reading is 250-400 wpm~\cite{doi:10.1177/1529100615623267}. In this work, the interactions times are estimated by counting the number of words in the narratives and considering the fastest pace for each channel: 160 wpm (auditory), and 400 wpm for (visual).\looseness=-1 \\
\textbf{Readability metric.} Since the narratives are generated using natural language, the Dale–Chall readability ${R_{DC}}$ formula~\cite{chall1995readability}, a well-known readability metric, is also used. Most of the readability metrics use a similar formula including two terms: (a) the proportion of `complex words' relative to the total number of words; and (b) the number of words per sentence. Usually, the word length or number of syllables is used to decide whether a text's words are `complex' (i.e. difficult to understand). More interestingly, Dale-Chall defines words as `complex' if they are not familiar (i.e. not included in a list of 3000 common words)~\cite{chall1995readability}. The resulting score indicates the reading level by educational grade needed to comprehend the text. Other readability metrics were considered and tried but results were similar, so we decided to use Dale-Chall because we consider `familiarity’ to be a more interesting indicator of complexity than `length’. 

\subsection{Results of the evaluation and discussion}
A statistical analysis was conducted to evaluate the significance of the proposed algorithms's improvement in reducing the amount of knowledge used in the explanations, and the explanation communication times, with respect to the baseline. We manipulated the two different algorithms for each specificity level (independent variable) and assessed them with respect to the metrics introduced in Section~\ref{sc:metrics} (dependent variables). Normality on the difference between the results obtained by both methods was assessed using the Shapiro-Wilk Test ($\alpha=0.05$). For the construction time, across all specificity levels, it is not possible to reject the normality assumption ($p>0.05$), while for the other metrics the normality assumption can be rejected ($p<0.05$). For the construction time and all three specificity levels, the results of the \textit{paired-t test} indicated that there is a significant large difference between the two algorithms, $p<0.05$. For the rest of metrics, the results of the \textit{Wilcoxon Signed Rank test} indicated a statistically significant difference in scores between the two algorithms for each of the specificity levels, $p<0.05$. The average evaluation metric results for the thirty pairs of plans, and each algorithm and level of specificity are summarized in Tab.~\ref{tab:average_results}. A summary of the specific statistical analysis results is provided in Tab.~\ref{tab:statistical_results}. More detailed results are available online.\footnote{\url{www.iri.upc.edu/groups/perception/ontology-based-explainable-robots}} 

{\cAlgorithmName} outperforms the baseline method in most cases, especially for levels two and three of specificity. Regarding the \textbf{number of tuples} $\mathcal{D_P}$, using {\cAlgorithmName} results in an overall reduction of more than 40\% and 70\% for levels 2 and 3 respectively. This is because {\cAlgorithmName} does a better selection of the narrative tuples. First, avoiding repeated tuples by collecting them for the whole pair instead of individually selecting tuples for each of the instantiated plans (see Sec.~\ref{sc:routine2}). Second, pruning the non-divergent knowledge between the plans (see Sec.~\ref{sc:routine3}). Hence, the contrastive narratives only contain what makes the plans different without undesired repetitions. For the \textbf{construction time} $C_T$, there are no major differences. However, it is worth commenting that the generation for level 3 would even require more than two seconds on average for any of the algorithms. Note that in some cases, the narrated plans contained more than 50 actions, hence, narratives of level 3 were really long. {\cAlgorithmName} produces a decrease in the \textbf{interaction times} ${I_T}_A$ and ${I_T}_V$ of approximately the 40\% and 70\% for specificity 2 and 3, respectively. The average times for the baseline method would be completely prohibitive for realistic interaction with humans. For level 3 of specificity, it would take 13 and 5 minutes for a human to listen and read the narratives, respectively. Indeed, although {\cAlgorithmName} reduces those times to 4 and 1.5 minutes, the improvement still falls short of ensuring a fluent and socially acceptable interaction. To overcome this, the robot might provide the short explanation of level 1, and only more details if required. Concerning this, {\cAlgorithmName} produces longer times when the specificity is 1. This is because {\cAlgorithmName} is more informative than the baseline method, since it includes the relationships between the plans at that level (e.g. shorter, better plan, etc.). Meanwhile, the baseline algorithm only states that both plans are instances of the class \textit{`Plan'}, refer to Olivares-Alarcos et al.~\cite{10161359} for more details about the baseline. Finally, in relation to the \textbf{readability index} ${R_{DC}}$ both behave similarly with a metric value close to 9, denoting a high portion of complex words. Specifically, such a value indicates that the explanations would be easily understood by an average college student. Hence, people with a lower education level may require a higher effort to interpret the narratives. Interestingly, the baseline method obtains a better value for specificity 3, because its narratives contain multiple short sentences, which is favored in the metric. In {\cAlgorithmName}, since the narratives are contrastive, they contain several subordinate sentences, which results in a higher degree of complexity.\looseness=-1

\begin{table}[t]
	\caption{Average evaluation results for each metric and the the 30 pairs of plans. The differences are statistically significant.}
	\label{tab:average_results}
    \centering
	\begin{tabular}{c|ccc|ccc|ccc}\toprule
		\textit{\textbf{Method}}&\multicolumn{3}{c}{\textbf{Baseline} (AXON)}& \multicolumn{3}{|c}{\textbf{ACXON}} & \multicolumn{3}{|c}{\textbf{p$<$0.05}}\\ 
		\textit{Specificity} &1&2&3 & 1&2&3 & 1&2&3  \\ \midrule
 \textit{$\mathcal{D_P}$} &\textbf{2.00}& 77.40& 305.73& 3.80& \textbf{44.27}&\textbf{91.33} & \checkmark& \checkmark&\checkmark\\
 \textit{$C_T$ (s)}& 0.73& 0.87& 3.50& \textbf{0.56}& \textbf{0.75}& \textbf{2.53} & \checkmark& \checkmark&\checkmark\\ 
  \textit{${I_T}_A$ (s)}&\textbf{5.25}&170.55& 782.80& 8.68& \textbf{101.00}&\textbf{258.85} & \checkmark& \checkmark&\checkmark\\ 
		\textit{${I_T}_V$ (s)}&\textbf{2.10}&68.22& 313.12& 3.47&\textbf{40.40}& \textbf{103.54} & \checkmark& \checkmark&\checkmark\\ 
 \textit{${R_{DC}}$} &9.62& \textbf{8.86}& \textbf{2.81}& \textbf{8.78}& 9.24&7.88 & \checkmark & \checkmark & \checkmark\\ 
	\end{tabular}
\end{table}

\begin{table}[t]
	\caption{Statistical evaluation results for each metric and specificity level. The specific statistical test metrics for each case is denoted between square brackets.}
	\label{tab:statistical_results}
    \centering
	\begin{tabular}{c|ccc}\toprule
		\textit{Specificity} &1&2&3 \\ \midrule
 \textit{$\mathcal{D_P}$ $[W,p,r]$} & $[0,< .001, 1.1]$ & $[0,< .001, -1.1]$ & $[0,< .001, -1.1]$ \\
 \textit{$C_T$ $[t(29),p]$}& $[84.9, <.001]$ & $[36.5, < .001]$ & $[21, < .001]$ \\ 
  \textit{${I_T}_A$ $[W,p,r]$}& $[0,< .001, 1.1]$ & $[0,< .001, -1.1]$ & $[0,< .001, -1.1]$ \\ 
		\textit{${I_T}_V$ $[W,p,r]$}& $[0,< .001, 1.1]$ & $[0,< .001, -1.1]$ & $[0,< .001, -1.1]$ \\ 
 \textit{${R_{DC}}$ $[W,p,r]$} & $[0, < .001, -0.9]$ & $[0, .046, 0.4]$ & $[0,< .001, 1.1]$ \\ 
	\end{tabular}
\end{table}

\section{May explanations be more selective?}
\label{sc:more_selective_explanations}
The evaluation results demonstrated that {\cAlgorithmName} enhances the selection of knowledge for explanatory contrastive narratives with respect to the baseline. However, it still requires long communication times for specificity levels 2 and 3, thus it shall be more selective. For instance, one might use the structure of the knowledge to constrain the tuples retrieval explained in Sec.~\ref{sc:routine2}, focusing only on part of the contrastive knowledge (e.g. only the plans' qualities). 

Following this rationale, it is proposed here a modification to the \textit{retrieve narrative tuples} routine to focus on specific aspects of plans. Specifically, at the second level of specificity, when the routine adds all the tuples in which at least one of the instances $\langle e_a, e_b \rangle$ is related through a property $p$ to an object $o$. Instead of considering tuples containing any object, the ontological class $c$ of the object is restricted: $\exists p \in N_P \land \exists c \in N_C \land (\langle e_a, p, o, t_i, t_f, sign \rangle \in \mathcal{G_T} \lor \langle e_b, p, o, t_i, t_f, sign \rangle \in \mathcal{G_T}) \land \langle o, type, c, t_i, t_f, sign \rangle$. This minor modification reduces the number of tuples retrieved at level 2, but its effect is also propagated to level 3, producing a larger decrease in the final number. Let's imagine that in the ongoing example from before, the algorithm is asked to compare the plans only using the qualities of plans (i.e. instances of \textit{dul.Quality}), the retrieved tuples would be reduced from 15 to 7: 

\vspace{2pt}
\noindent\fbox{%
    \parbox{0.96\columnwidth}{%
        {\footnotesize
        $\boldsymbol{\mathcal{T_P}_{1}}$ $\langle$`bringing tea', isBetterPlanThan, `bringing cola', \_, Inf, positive$\rangle$, \\
        $\boldsymbol{\mathcal{T_P}_{2}}$ $\langle$`bringing tea', isCheaperPlanThan, `bringing cola', \_, Inf, positive$\rangle$, \\
        $\boldsymbol{\mathcal{T_P}_{3}}$ $\langle$`bringing cola', hasCost, `cola cost', \_, Inf, positive$\rangle$, \\
        $\boldsymbol{\mathcal{T_P}_{4}}$ $\langle$`bringing tea', hasCost, `tea cost', \_, Inf, positive$\rangle$, \\
        $\boldsymbol{\mathcal{T_P}_{5}}$ $\langle$`cola cost', hasDataValue, `59', \_, Inf, positive$\rangle$, \\
        $\boldsymbol{\mathcal{T_P}_{6}}$ $\langle$`tea cost', hasDataValue, `27', \_, Inf, positive$\rangle$, \\
        $\boldsymbol{\mathcal{T_P}_{7}}$ $\langle$`cola cost', hasWorseQualityValueThan, `tea cost', \_, Inf, positive$\rangle$.
        }
    }%
}
\vspace{2pt}

The rest of algorithm's routines would be applied as it was shown before, producing a shorter narrative: 

\vspace{2pt}
\noindent\fbox{%
    \parbox{0.96\columnwidth}{%
        {\footnotesize
        `bringing tea' is better plan than and is cheaper plan than `bringing cola'. 
        `bringing cola' has cost `cola cost', which has a higher value than `tea cost'; while `bringing tea' has cost `tea cost'. 
        `cola cost' has value `59', while `tea cost' has value `27'. 
        }
    }%
}
\vspace{2pt}

With this modification {\cAlgorithmName} does, by construction, a more selective knowledge retrieval to build the contrastive narrative, which would shorten the explanations and reduce the communication time. Furthermore, the algorithm now allows to constrain the preferred content of the contrastive narratives, which might be used to provide personalized explanations. Hence, the modification, while being simple, it contributes to generate potentially more socially acceptable explanations.  

Note that there is a trade-off between brevity of an explanation and its informativeness, and this trade-off is conditioned by the preferences of the final user. Shortening explanations too much risks omitting relevant knowledge, which can lead to poor perceived quality, so how the selection is done is essential. From social sciences, we know that explanations are often selected in a biased way~\cite{MILLER20191}, so having the ability to be more selective is desirable. However, this makes the objective evaluation of the modified algorithm less insightful than in Section~\ref{sc:evaluation_general_ref}. To assess how a more selective strategy affects this balance, it would be sensible to conduct a user study with different selective explanations, as we plan to do in future work. Additionally, we aim to explore how user preferences might be identified or learned, through interaction, so that the algorithm can tailor the selection process accordingly. In cases where preferences are unknown, providing longer explanations might help prevent oversimplification.



\section{Conclusion}
\label{sc:conclusion}
This work presents a method for robots to model and reason about the differences between plans, to infer which one is better and to narrate the inferences to other agents (e.g. humans). The approach comprises a novel ontological model for robots to describe plans and their qualities for reasoning during plans comparison, and a new algorithm to construct ontology-based contrastive explanatory narratives. The approach is general to be used with other ontologies, beyond the case of contrasting plans (e.g. modeling the qualities of two drinks and narrating the differences). The model is validated by instantiating it to answer a set of competency questions. The algorithm is evaluated against a baseline with respect to a set of objective metrics. Our solution outperforms the baseline in general, doing a better selection of knowledge tuples to build the explanation (avoiding non-divergent knowledge), and producing explanations that would require less time for the robot to communicate them. It is also discussed a final refinement of the algorithm to be more selective, shortening the narratives and opening the door for more personalized explanations. In the future, we will look at making more accessible the narratives' language, and a user study will be conducted to evaluate their quality. Note that robots add the possibility of physically executing the plan, which opens issues to investigate: the ‘preferred’ moment to explain (e.g. before or after executing), or the context in which the competing plans are conceived (e.g. due to an adaptation while executing).\looseness=-1



\section*{Acknowledgments}
\small{
This work was partially supported by Project PID2023-152259OB-I00 funded by MCIN/ AEI /10.13039/501100011033; Project PCI2020-120718-2 funded by MCIN/ AEI /10.13039/501100011033 and by the "European Union NextGenerationEU/PRTR"; Project EP/V062506/1 funded by EPSRC (UK); and also by the European Union under the project ARISE (HORIZON-CL4-2023-DIGITAL-EMERGING-01-101135959). G. Canal was supported by the Royal Academy of Engineering and the Office of the Chief Science Adviser for National Security under the UK Intelligence Community Postdoctoral Research Fellowship programme. A. Olivares-Alarcos was supported by the European Social Fund and the Ministry of Business and Knowledge of Catalonia through the FI 2020 grant.
}

\bibliographystyle{elsarticle-num} 
\bibliography{sample.bib}

\end{document}